\documentclass{article}
\PassOptionsToPackage{numbers,sort&compress}{natbib}
\usepackage[final,sglblindworkshop]{neurips_2026}
\workshoptitle{Predictive AI Evaluation Competition (PAIEC)}
\usepackage[T1]{fontenc}
\usepackage[utf8]{inputenc}
\usepackage{hyperref}
\usepackage{url}
\usepackage{booktabs}
\usepackage{enumitem}
\usepackage{amsmath}
\usepackage{graphicx}
\usepackage{xcolor}
\usepackage{float}

\title{How Much Do Language Models Already Know About AI Systems on a Benchmark?}

\author{%
  Siwei Shan\thanks{Equal contribution.}\\
  NLPQuant AI Inc.\\
  \texttt{siwei@nlpquant.ai}
  \And
  Sunny Yang\footnotemark[1]\\
  University of Illinois Urbana-Champaign\\
  \texttt{guangy4@illinois.edu}
  \And
  Weiyan Zhao\\
  NLPQuant AI Inc.\\
  \texttt{weiyan@nlpquant.ai}
}

\begin{document}

\maketitle

\begin{abstract}
We predict if an AI system will answer each question of a benchmark correctly, starting before we see any of its graded answers. Claude reads the name of the system, a few public facts and a few sample questions, and estimates its accuracy. On seven public benchmarks not used in development, each with at least 25 systems, the estimates explain 50 to 89\% of the true differences in accuracy between systems. This share shows the builders of a benchmark how much of the differences between known systems a language model can already guess. The prompt has no results of these benchmarks, but the benchmarks and the systems all predate the training cutoff of Claude Opus 5.5. On eight benchmarks released after the cutoff, estimates without sample questions order systems little better than their release dates. Before the first graded answer, the estimate is about as accurate as four to eight graded answers. In simulation, it gives most of the gain of the language-model parts of our submitted predictor. In trial runs of the Predictive AI Evaluation Competition at NeurIPS 2026, it made the Brier score on hidden benchmarks before the first answer about a fifth lower than a 50\% guess. We release a script that computes this share for other benchmarks with at least 25 measured systems.
\end{abstract}

\begin{figure}[h]
\centering
\includegraphics{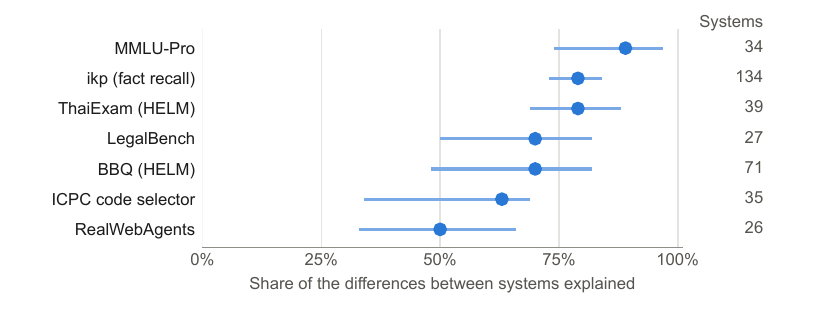}
\caption{\textbf{With no results of these benchmarks in the prompt, Claude's estimates explain 50 to 89\% of the true differences between systems.} Each row is a benchmark with at least 25 systems released before March 2026. The systems and the benchmarks are all older than the training cutoff of Claude. The line is the 90\% interval.}
\label{fig:known}
\end{figure}

\section{Introduction}
\label{sec:intro}

The Predictive AI Evaluation Competition asks for the probability that an AI system answers each question of an unseen benchmark correctly. The system reveals its graded answers one at a time. The competition scores the predictions after 0, 1, 3, 7, 15 and 31 answers.

The score uses the Brier score, which is the squared error of a predicted probability. The final score is the area under this learning curve (ALC). The budgets of 0, 1 and 3 answers carry half of its weight, and the error is largest there.

With no answers yet, a predictor cannot see the difference between a system that answers almost all questions correctly and one that answers half.

\begin{figure}[t]
\centering
\includegraphics{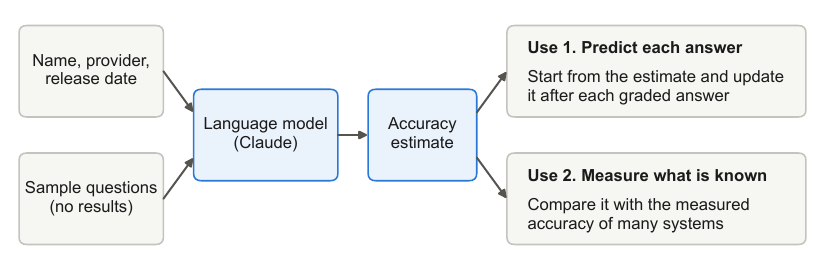}
\caption{\textbf{A language model turns a system's name and a few questions into an accuracy estimate with two uses.} The inputs contain no results of the target case.}
\label{fig:method}
\end{figure}

Our idea is to ask a language model. A language model can estimate how often a named system will succeed before the system answers anything. We make a prior from this estimate, and the answers of the system then update the prior (Figure~\ref{fig:method}).

We call a system under a given configuration a subject, and one subject on one benchmark a case. These are our contributions.
\begin{itemize}
\item \textbf{The finding} (Section~\ref{sec:prior}). Language models know how strong a named AI system is. They order 77 to 80\% of the pairs of systems on a benchmark correctly. An order by release date, newest first, gets 64\%.
\item \textbf{Where it comes from.} For Claude, with the same sample questions in the prompt, the name of the system carries almost all of this knowledge. On the public score table of Epoch AI, systems released after the training cutoff are ordered less well, but still better than by release date. A regression on name features, without a language model, loses about as much, so newer systems are also harder to order.
\item \textbf{The effect} (Sections~\ref{sec:ablate} and~\ref{sec:online}). A predictor that starts from the estimate gains most before the first answer in simulation, and the estimate also lowered the error before the first answer in the trial runs of the competition.
\item \textbf{A reusable kit.} We release the predictor, a simulator of the competition, the cached Claude outputs and a script that measures how much of a benchmark a language model knows. We also release 1,514 accuracy forecasts, 757 on competition benchmarks and 757 on new benchmarks, on which other methods can be scored.
\end{itemize}

\paragraph{Related work.} The Bayesian Linguistic Evaluator of the organisers~\cite{paiecbaseline} asks a language model about each response separately. We ask for each case, not for each response, and the graded answers then update a Beta prior. Park et al.\ ask language models to forecast benchmark scores from task descriptions \cite{park2025}. Zeng and Papailiopoulos complete the missing scores of a model from its other scores, and a language model does better there with real names than with anonymised names \cite{zeng2026}.

Methods that predict from a few questions usually need questions that are already calibrated on many models \cite{polo2024tiny,hofmann2025,vivek2023}, or they learn the difficulty of a question from its text \cite{truong2025}. In the same competition, Ergun et al.\ start from a weak Beta prior that they choose from public counts \cite{ergun2026}.

\section{Task}
\label{sec:task}

For each case, the competition splits the questions 50/50 into an acquisition pool and an evaluation pool. The acquisition questions come to the predictor as a stream. The predictor decides which of these questions to reveal, that is, for which questions it gets the graded answer. It can reveal up to 31. The competition scores the predictions on the evaluation pool at budgets $b \in \{0,1,3,7,15,31\}$ with weights $0.1, 0.2, 0.2, 0.2, 0.2, 0.1$, and averages the ALC over the cases.

The predictor sees the revealed answers of every case in a run, each marked correct or wrong. It also sees the question text and the attributes of the subject. It never sees the reference answers.

\section{Method}
\label{sec:method}

\paragraph{The estimate.} We ask Claude Opus~5.5 what fraction of the questions of the benchmark the subject answers correctly. The prompt has two parts.
\begin{itemize}[leftmargin=*,itemsep=0pt,topsep=2pt]
\item The target. Up to eight sample questions of the benchmark (five for the estimates of Figures~\ref{fig:known}, \ref{fig:worth} and~\ref{fig:order}b), and the name, provider, release date, harness and reasoning effort of the subject.
\item The scale. Eight calibration examples, each with questions of a public benchmark, a system and its measured accuracy.
\end{itemize}
Eight competition benchmarks supply one calibration example each. Only for these eight benchmarks does the prompt contain a result of the target benchmark. The exact prompt is \texttt{PRIOR\_PROMPT} in \texttt{submission/model.py} of the release.

\paragraph{The update.} The prediction is a weighted average of the estimate and the graded answers, where the estimate counts as four answers. We first shrink the estimate $q_s$ toward one half, $m_s = 0.5 + 0.75\,(q_s - 0.5)$. After the subject has revealed the set $R_s$ of questions, with outcomes $y_{sj}$ equal to 1 if correct and 0 if wrong, the prediction is
\begin{equation}
\hat\mu_s = \frac{\sum_{j \in R_s} y_{sj} + 4\, m_s}{|R_s| + 4}.
\end{equation}

\paragraph{The rest of the predictor.} Four more parts set the prediction (Appendix~\ref{app:predictor}).
\begin{enumerate}[leftmargin=*,itemsep=0pt,topsep=2pt]
\item Item ratings. The language model estimates how hard each question is for a typical system.
\item Other subjects. The answers of other subjects on the same benchmark.
\item Similar questions. The answers to questions with similar text.
\item Identical questions. The answers of other subjects to the same question.
\end{enumerate}
A fifth part chooses which questions to reveal.

\section{The language model knows the system}
\label{sec:prior}

\paragraph{How we score the estimate.} We count how many of its own graded answers a subject needs to match the language model. In other words, $n^*$ is the number of own answers with the same expected squared error. With $n$ random answers, $k$ of them correct, a simple estimate is $(k+1)/(n+2)$. For the estimate $m$ and the true accuracy $a$, the language model is worth
\begin{equation}
n^* = \min\Bigl\{n : \mathrm{E}\Bigl[\bigl(\tfrac{k+1}{n+2} - a\bigr)^2\Bigr] \le \mathrm{E}\bigl[(m - a)^2\bigr]\Bigr\}
\end{equation}
answers, where $k$ is binomial and the expectations run over the cases of a set.

\begin{figure}[t]
\centering
\includegraphics{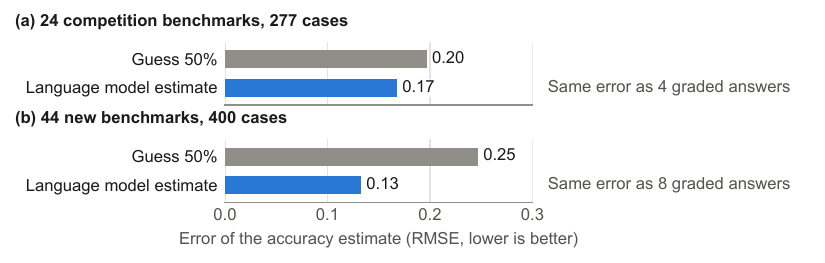}
\caption{\textbf{Before any answers, the estimate is as accurate as four to eight of the system's own graded answers.} The bars show the error of the accuracy estimate. Lower is better.}
\label{fig:worth}
\end{figure}

\paragraph{Two sets of cases.} We use two sets of public cases (Table~\ref{tab:sets}). The competition cases are the 277 cases of the 24 competition benchmarks that supply no calibration example. We used most of these benchmarks in development. The new set comes from 49 benchmarks that we downloaded after the final submission. Its first run has 400 cases from 44 of them, and a second run added 357 more.

\paragraph{The estimate is worth four to eight answers.} On the 277 competition cases, the RMSE of the estimate is 0.17, against 0.20 for a guess of 50\%. This is worth four answers (Figure~\ref{fig:worth}a). On the 400 new cases of the first run, the RMSE is 0.13, against 0.25, and the estimate is worth eight answers (Figure~\ref{fig:worth}b).

Two priors use data alone, the mean accuracy on other benchmarks and the level of the subject on other benchmarks. On the 277 competition cases, both are worth no answers (Table~\ref{tab:prior_full}).

\paragraph{The share known in advance.} This share is the part of the true differences between systems that the estimates explain. On one benchmark, let $e$ be the estimate errors after we remove their mean, $a$ the true accuracies and $r$ the split-half reliability of the accuracies. The share is
\begin{equation}
K = 1 - \frac{\operatorname{Var}(e)}{r\,\operatorname{Var}(a)}.
\end{equation}
The factor $r$ removes the noise from the finite number of questions from the differences between systems. The same noise also enters the errors, so $K$ is, if anything, too low. Because the mean error is removed, $K$ scores the order and spacing of the systems, not their level.

On seven benchmarks with at least 25 systems, the share is 50 to 89\% (Figure~\ref{fig:known}). All these systems and all seven benchmarks (released from October 2021 to April 2026) are older than the training cutoff of Claude Opus 5.5 (June 2026). The script \texttt{tools/known\_share.py} in the release computes the share for other benchmarks.

\begin{figure}[t]
\centering
\includegraphics{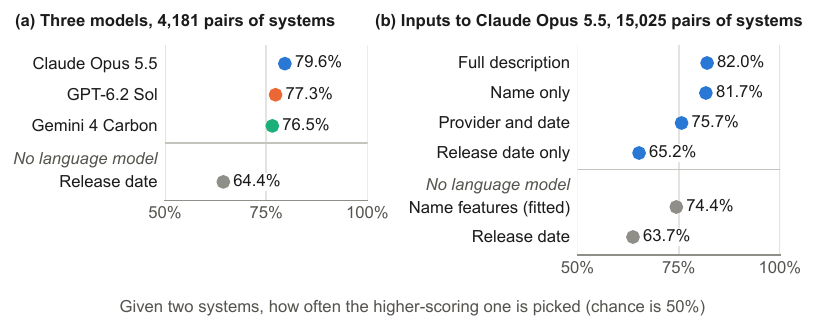}
\caption{\textbf{Language models order most pairs of systems correctly, and for Claude, with the same sample questions, the name of the system gives almost all of this.} The two panels use different sets of cases, so compare rows only within a panel. Gray rows use no language model.}
\label{fig:order}
\end{figure}

\paragraph{Other language models also know the system.} We gave the same prompt with eight sample questions to GPT-6.2 Sol, Gemini~4 Carbon and Claude Opus~5.5. We used 357 of the 400 first-run cases, those with systems released before the cutoff of all three models. These cases give 4,181 pairs of systems. Each model orders 77 to 80\% of them correctly, against 64\% for the order by release date (Figure~\ref{fig:order}a).

\paragraph{The knowledge comes from the name.} With only the name of the system, Claude orders 81.7\% of the pairs correctly, against 82.0\% with the full description (Figure~\ref{fig:order}b). The difference is under half a point (90\% interval $-0.6$ to $+1.1$). This test uses the 666 cases of both runs with an exact release date before March 2026, and every variant sees the same five sample questions. On the 71 cases of Table~\ref{tab:prior_full}, without the name, provider and release date, the estimate is almost as poor as a constant.

\paragraph{Naming the benchmark changes little.} The deployed prompt never names the benchmark, but its sample questions may reveal it. When we name the benchmark, the error barely changes (Table~\ref{tab:prior_full}). This test cannot tell whether Claude recalls results of the benchmark. On benchmarks released after its training cutoff, Opus, with a prompt that names the benchmark, does not clearly order systems better than the release date (Section~\ref{sec:limits}).

\paragraph{Claude does more than read the name.} We fitted a regression on features of each system. They are the parameter count, tier words such as mini or pro and the version from the name, and the provider and release date. Unlike Claude, this regression learns from the scores of other systems. It orders 74.4\% of the pairs correctly. Claude with only the name orders 7 points more (90\% interval 6 to 13).

\paragraph{The advantage fades after the training cutoff.} To test this, we used an older model, GPT-4.1, whose training cutoff is June 2024. Many systems in our data came later. On the competition and new benchmarks, it orders 21 points more pairs correctly than the release date before its cutoff, and 4 points fewer after it. Here each benchmark counts once (Table~\ref{tab:cutoff}).

\begin{figure}[t]
\centering
\includegraphics{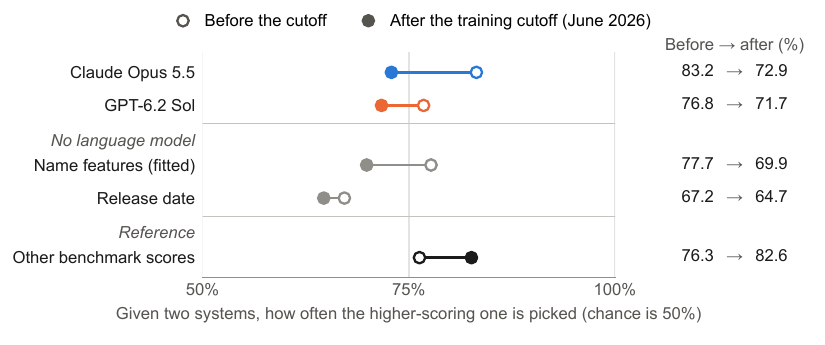}
\caption{\textbf{For systems released after the training cutoff, the language models order fewer pairs correctly, but more than the release date.} The regression on name features drops about as much, so newer systems are also harder to order from their names, providers and release dates. The bottom row uses published scores on other benchmarks.}
\label{fig:cutoff}
\end{figure}

\paragraph{Newer systems are harder to order, with or without a language model.} The public score table of Epoch AI \cite{epoch2026} lists 44 systems released after the cutoff of Opus and GPT-6.2 Sol. For this table, the prompt names the benchmark and shows no sample questions. For the newer systems, Opus orders fewer pairs correctly, and the regression on name features drops about as much (Figure~\ref{fig:cutoff}). Opus still orders 8 points more pairs correctly than the release date (90\% interval 5 to 12, 26 benchmarks).

Published scores on other benchmarks, which the language models do not see, order the newer systems best (83\%).

\paragraph{More than a general level.} The scores of a model across benchmarks are low-dimensional \cite{ruan2024,zeng2026}. So we predicted the accuracy of each system on a benchmark from two general factors of the system, fitted on its scores on other benchmarks. Adding the estimate of the language model cuts the squared error of this prediction by 42\% (90\% interval 29 to 52, on 383 cases of the new set).

\paragraph{Systems, not questions.} On four of five large benchmarks, the estimates of Opus explain 69 to 91\% of the differences between systems (squared correlation, corrected for reliability), but its ratings explain only 7 to 42\% of the differences between questions (Figure~\ref{fig:sysq}). The exception is \texttt{ikp}, a benchmark of fact-recall questions.

\section{The prior improves the predictor}
\label{sec:ablate}

\paragraph{The test.} We turned the two language-model parts on and off, the subject prior and the item ratings. A draw is one simulated competition run. We ran the submitted predictor on 90 draws from the public benchmarks that we used in development. Each draw has 3 to 8 cases, mostly with one subject per benchmark. All variants use the same draws and the same cached Opus estimates.

\begin{table}[t]
\centering
\caption{\textbf{The accuracy prior gives most of the gain of the two language-model parts of the submitted predictor.} ALC is the area under the Brier learning curve, and $B_0$ is the Brier score before the first answer. Lower is better.}
\label{tab:ablate}
\small
\begin{tabular}{lcccc}
\toprule
Variant & ALC & $B_0$ & ALC difference & Better in \\
\midrule
Laplace estimate on own answers & 0.2020 & 0.250 & $+0.0008 \pm 0.0002$ & 20\% of draws \\
Neither part & 0.2012 & 0.250 & -- & -- \\
Item ratings only & 0.1992 & 0.244 & $-0.0020 \pm 0.0004$ & 68\% of draws \\
Subject prior only & 0.1941 & 0.221 & $-0.0071 \pm 0.0009$ & 80\% of draws \\
Both (submitted) & \textbf{0.1924} & 0.216 & $-0.0088 \pm 0.0010$ & 83\% of draws \\
\bottomrule
\end{tabular}
\end{table}

\paragraph{The prior gives most of the gain.} The subject prior lowers the Brier score before the first answer from 0.250 to 0.221 (Table~\ref{tab:ablate}). Each difference in the table is against the variant with neither part, paired over draws, with one standard error.

\begin{figure}[t]
\centering
\includegraphics{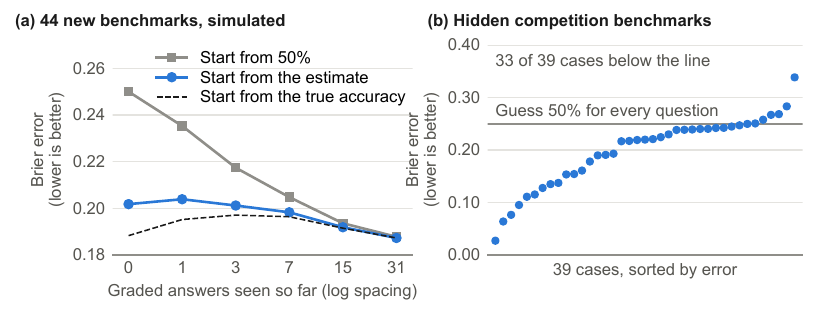}
\caption{\textbf{The estimate helps most before the first answer in simulation, and it also helped before the first answer on the hidden competition benchmarks.} (a) 30 simulated draws on the new benchmarks, with item ratings off. Each line is the submitted predictor with a different start. The 50\% start and the estimate also enter with different weights (Table~\ref{tab:parts}). (b) Each dot is one hidden case before any graded answer.}
\label{fig:pred}
\end{figure}

\paragraph{The effect fades after about seven answers.} Figure~\ref{fig:pred}a starts the same predictor from 50\%, from the estimate, and from the true accuracy of the case. With a good start, the error first increases, because the predictor gives the first answers too much weight. After about seven answers, the three starts give almost the same error.

\section{Live results}
\label{sec:online}

During the competition, a formative run (a trial run) scored a submission on eight or nine cases from seven hidden benchmarks and gave feedback. Before the first answer, a predictor without either language-model part predicts one half for every question, with a Brier score of 0.250.

Five runs used the prior without item ratings. Pooled over these runs, the Brier score before the first answer was 0.199 on 39 distinct cases (90\% interval 0.166 to 0.227). The prior lowered the error on 33 of these 39 cases (Figure~\ref{fig:pred}b), and, averaged per benchmark, on six of the seven benchmarks. Appendix~\ref{app:runs} lists every run.

\section{Limitations}
\label{sec:limits}

\begin{enumerate}
\item \textbf{Unknown systems.} The prior needs a system that the language model knows, and the predictor does not use it for unnamed systems. Of 15 held-out benchmarks that we never used in development, 8 have named subjects. On these 8, the prior made no measurable difference (ALC change $-0.0011$, 90\% interval $-0.009$ to $+0.007$, Appendix~\ref{app:heldout}).
\item \textbf{Systems after the cutoff.} On Epoch's table, Opus orders the after-cutoff systems only 3 points better than the regression on name features, and the 90\% interval reaches zero. Its gain over the release date comes from ranking the Anthropic and OpenAI systems against those of other developers. Among the Anthropic and OpenAI systems alone, and among the other systems alone, it is no better than the release date.
\item \textbf{Benchmarks released after the cutoff.} On eight benchmarks of Epoch's table released after the training cutoff of Opus, its estimates still explain 49\% of the variance between systems (squared correlation). But the release date alone explains 35\%, and Opus orders only 2 points more pairs correctly than the release date (90\% interval $-1$ to $+6$). We made this check after the main analysis.
\item \textbf{Other models after the cutoff.} GPT-4o and GPT-4.1 did not clearly beat the release date after their cutoffs (Appendix~\ref{app:more}). For Opus on Epoch's table, we used a short neutral system prompt. GPT-6.2 Sol ran with its default system prompt, and Gemini was not tested there.
\item \textbf{Analyses after a first look.} We chose these after we saw first results. They are the share known in advance (only the offset removed) and its threshold of 25 systems, the split by developer, the neutral rerun before the cutoff, the comparisons of Opus with the regression on name features on Epoch's table, the comparison of the name with the full description, and the rerun of the systems-versus-questions comparison (Figure~\ref{fig:sysq}) on denser data after the first attempt on sparse data did not meet its rule. Because of this, we do not rank benchmarks by the share.
\item \textbf{Weighting.} The ordering numbers of Figure~\ref{fig:order} count each pair of systems once. If each benchmark counts once instead, the three models order 71 to 75\% of the pairs correctly, and the release date 63\%.
\item \textbf{Uncertainty.} On the competition cases, the 90\% interval of ``worth four answers'' is 0 to 10, so this number is rough.
\item \textbf{A weak baseline.} The two-factor baseline is weak on our sparse data. It explains only 22\% of the differences between systems.
\item \textbf{Development data.} We used the benchmarks of Table~\ref{tab:ablate} in development, so its gains may be optimistic. On one of the seven hidden benchmarks, the prior made the error worse ($+0.018$).
\end{enumerate}

Appendix~\ref{sec:repro} gives the code, the reproduction steps and the disclosures.

\bibliographystyle{plain}

\appendix

\section{Reproducibility and disclosures}
\label{sec:repro}

\begin{figure}[H]
\centering
\includegraphics{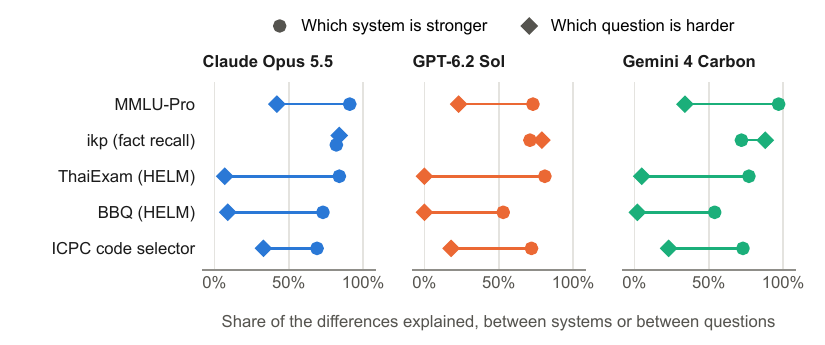}
\caption{\textbf{Language models know which system is stronger far better than which question is harder, except on ikp.} Circles are for systems and diamonds for questions. This figure uses the squared correlation corrected for reliability, not the measure of Figure~\ref{fig:known}.}
\label{fig:sysq}
\end{figure}

The code, the simulator, the cached Claude outputs and the simulation logs are at \url{https://github.com/sunnyadn/paiec-pooled-priors}. They cover Figure~\ref{fig:worth} and Tables~\ref{tab:ablate}, \ref{tab:prior_full} and~\ref{tab:heldout}, and they hold the estimates behind Figure~\ref{fig:known}. The script \texttt{tools/known\_share.py} computes its shares, but its input files of true accuracies need measurement-db and are not built by a released script. The true-accuracy start of Figure~\ref{fig:pred}a is not in the release. Figure~\ref{fig:pred} also needs the gated data and the private logs of the competition. Apart from a few development comparisons (\texttt{data/llm/prior\_eval\_x*.json} and \texttt{data/llm/prior\_cache\_gpt.jsonl}), the outputs of GPT and Gemini, and the analyses behind Figures~\ref{fig:order}, \ref{fig:cutoff} and~\ref{fig:sysq} and Tables~\ref{tab:factors} and~\ref{tab:cutoff} are not released. The command \texttt{python reproduce.py} prints the simulation tables from the logs. The command \texttt{python reproduce.py -{}-rerun} reruns the ablation and held-out simulations from stored draw identifiers, without a model API key but with access to measurement-db.

With access to the gated measurement-db, the documented scripts rebuild the question files. After that, \texttt{reproduce.py} and \texttt{src/score\_forecasts.py} recompute the errors of Figure~\ref{fig:worth} and Table~\ref{tab:prior_full}, and \texttt{src/sim\_extra.py} reruns the simulations on the 44 new benchmarks.

The final Codabench submission is 1017682. It needs only Python and NumPy. Without an API key, it skips the language-model parts.

\paragraph{Disclosures.} We used the public measurement-db to develop the predictor, to simulate the competition and to make the eight solved examples. At test time, the predictor called Claude Opus~5.5 through the Anthropic API, with Claude Sonnet~5.5 and Haiku~5.5 as fallbacks for the subject prior. During development, we also compared GPT and Gemini models, and we used the coding assistants Claude Code and Codex. Table~\ref{tab:access} gives how we called the language models for the comparisons it lists.

\begin{table}[H]
\centering
\caption{How we called the language models after the competition.}
\label{tab:access}
\small
\begin{tabular}{lll}
\toprule
Result & Model & Access \\
\midrule
Figure~\ref{fig:order}a & Claude Opus~5.5 & Claude Code CLI \\
 & GPT-6.2 Sol & OpenAI API (no number on 1 of 677 forecasts) \\
 & Gemini~4 Carbon & Google API, 149 forecasts via the Arena API (no number on 3) \\
Figure~\ref{fig:cutoff} & Claude Opus~5.5 & Claude Code CLI, short neutral system prompt \\
 & GPT-6.2 Sol & Codex CLI, no web search \\
Figure~\ref{fig:sysq} & Gemini~4 Carbon & item ratings and about half of the system forecasts via the Arena API \\
\bottomrule
\end{tabular}
\end{table}

The 677 forecasts are the 277 competition cases and the 400 new cases of the first run. Figure~\ref{fig:order}a uses a separate run of Opus with eight sample questions on 357 of the first-run cases.

\section{More results on the knowledge of the language models}
\label{app:more}

\begin{table}[H]
\centering
\caption{The estimate adds to a two-factor model of each system. We fit two general factors per system on its other benchmarks and predict its accuracy on the target benchmark. The cells give the share of the remaining squared error that adding the estimate removes. Brackets are 90\% intervals. The case counts are for Opus. GPT-6.2 Sol and Gemini~4 Carbon have about 250 new and 98 to 104 competition cases. With factors from our sparse data, the two-factor model alone explains 22\% of the differences between systems on the new benchmarks. With factors from Epoch's table, it explains 39\%. The GPT and Gemini columns were added after the main analysis.}
\label{tab:factors}
\small
\begin{tabular}{llccc}
\toprule
Data & Cases & Opus & GPT-6.2 Sol & Gemini~4 Carbon \\
\midrule
New benchmarks & 383 & 42\% [29, 52] & 38\% & 43\% \\
Competition benchmarks & 104 & 30\% [16, 47] & 23\% & 28\% \\
New, factors from Epoch's table & 308 & 32\% [16, 48] & -- & -- \\
\bottomrule
\end{tabular}
\end{table}

\begin{table}[H]
\centering
\caption{Points of pairs of systems ordered correctly above the order by release date, before and after the training cutoff of each estimator. The GPT-4o and GPT-4.1 rows use benchmarks with at least four systems in each window. For GPT-4o, the size of the drop with each benchmark counted once depends on the minimum number of systems per benchmark. In a pilot on 13 cases with systems released from June 2026, the RMSE of Opus was 0.30, against 0.32 for a guess of 50\%.}
\label{tab:cutoff}
\small
\begin{tabular}{llcc}
\toprule
Estimator & Data and weighting & Before & After \\
\midrule
GPT-4o (cutoff Oct.\ 2023) & competition and new, each pair once & $+16$ & $-7$ \\
GPT-4.1 (cutoff June 2024) & competition and new, each pair once & $+14$ & $+3$ \\
GPT-4.1 (cutoff June 2024) & competition and new, each benchmark once & $+21$ & $-4$ \\
Opus 5.5, neutral system prompt & Epoch, each benchmark once & $+16$ & $+8$ \\
Opus 5.5, default system prompt & Epoch, each benchmark once & $+14$ & $+10$ \\
GPT-6.2 Sol, default prompt & Epoch, each benchmark once & $+10$ & $+7$ \\
\bottomrule
\end{tabular}
\end{table}

\section{The quality of the prior in detail}
\label{app:prior}

\begin{table}[H]
\centering
\caption{The sets of public cases. The new set has a first run of 400 cases and a second run of 357 more. Figure~\ref{fig:order}a uses a different group of 357 cases, the first-run cases with systems released before the training cutoff of all three models.}
\label{tab:sets}
\small
\begin{tabular}{lcccl}
\toprule
Set & Benchmarks & Cases & In development & Used in \\
\midrule
Development & all used & -- & yes & Table~\ref{tab:ablate} \\
Competition, no calibration example & 24 & 277 & mostly & Figure~\ref{fig:worth}a \\
Competition, with a calibration example & 8 & 480 & yes & -- \\
Held-out & 15 & -- & no & Appendix~\ref{app:heldout} \\
New, first run & 44 & 400 & no & Figures~\ref{fig:worth}b, \ref{fig:order}a, \ref{fig:pred}a \\
New, both runs & 49 & 757 & no & Figures~\ref{fig:known}, \ref{fig:order}b \\
\bottomrule
\end{tabular}
\end{table}

\begin{table}[H]
\centering
\caption{The error of the accuracy estimate before any answers. ``Answers'' is the number of the subject's own answers with the same error (Section~\ref{sec:prior}). The lower rows use an earlier sample of 71 cases on which all models gave a number. $^*$This row uses answers of other subjects, which do not exist before the first answer.}
\label{tab:prior_full}
\small
\setlength{\tabcolsep}{4pt}
\begin{tabular}{lcccc}
\toprule
Estimate & Cases & RMSE & Correlation & Answers \\
\midrule
Constant 0.5 & 277 & 0.197 & -- & 0 \\
Mean accuracy of public benchmarks, this one left out & 277 & 0.199 & -- & 0 \\
Subject's level from its other public benchmarks & 277 & 0.203 & -- & 0 \\
Deployed: Opus, then Sonnet, then Haiku & 277 & \textbf{0.168} & 0.55 & \textbf{4} \\
Subject's level plus this benchmark's level from other subjects$^*$ & 277 & 0.143 & -- & 7 \\
\midrule
Constant 0.5 & 71 & 0.217 & -- & 0 \\
Claude Haiku~5.5 & 71 & 0.185 & 0.63 & 3 \\
Claude Sonnet~5.5 & 71 & 0.142 & 0.76 & 7 \\
Claude Opus~5.5 & 71 & 0.141 & 0.76 & 7 \\
Opus, subject's identity hidden & 71 & 0.211 & 0.28 & 1 \\
Opus, benchmark named in the prompt & 71 & 0.136 & 0.78 & 8 \\
\bottomrule
\end{tabular}
\end{table}

\section{Benchmarks never used in development}
\label{app:heldout}

\begin{table}[H]
\centering
\caption{The change in Brier score from the subject prior on the eight held-out benchmarks with named subjects, over 90 simulated draws, with item ratings off. Negative is better. The prior helps where its estimate is closer to the true accuracy than 0.5 is. On the seven held-out benchmarks with mostly unnamed subjects, the predictor does not use the prior.}
\label{tab:heldout}
\small
\begin{tabular}{lrrrrrr}
\toprule
Benchmark & Case draws & $B_0$ & $B_1$ & $B_3$ & $B_7$ & ALC \\
\midrule
\texttt{bertaqa} & 27 & $-0.059$ & $-0.041$ & $-0.023$ & $-0.009$ & $-0.021$ \\
\texttt{evouna} & 54 & $-0.050$ & $-0.036$ & $-0.012$ & $-0.002$ & $-0.015$ \\
\texttt{aider} & 49 & $-0.029$ & $-0.025$ & $-0.018$ & $-0.004$ & $-0.013$ \\
\texttt{healthadminbench} & 47 & $-0.035$ & $-0.018$ & $-0.006$ & $-0.000$ & $-0.008$ \\
\texttt{edu\_circuit\_hw} & 46 & $-0.015$ & $-0.010$ & $-0.007$ & $-0.003$ & $-0.005$ \\
\texttt{clasheval} & 51 & $+0.017$ & $+0.009$ & $+0.006$ & $+0.000$ & $+0.005$ \\
\texttt{dbpa} & 59 & $+0.051$ & $+0.027$ & $+0.013$ & $+0.003$ & $+0.014$ \\
\texttt{ehrflowbench} & 51 & $+0.056$ & $+0.035$ & $+0.025$ & $+0.012$ & $+0.023$ \\
\midrule
All eight & 384 & $-0.003$ & & & & $-0.001$ \\
\bottomrule
\end{tabular}
\end{table}

\section{Online runs}
\label{app:runs}

\begin{table}[H]
\centering
\caption{Every online formative run. Section~\ref{sec:online} pools the five runs with the subject prior and without item ratings, with a bootstrap over the seven hidden benchmarks. $B_0$ and $B_{31}$ are the mean Brier scores at budgets 0 and 31. Excess is the ALC above the level at budget 31. The runs drew different subjects, so the rows are not a paired comparison.}
\small
\begin{tabular}{llccccc}
\toprule
Submission & Version & Cases & $B_0$ & Raw ALC & $B_{31}$ & Excess \\
\midrule
1016627 & v6 & 9 & 0.250 & 0.158 & 0.116 & 0.042 \\
1016628 & v6 & 9 & 0.250 & 0.170 & 0.143 & 0.027 \\
1016651 & v6, no API key & 8 & 0.250 & 0.158 & 0.134 & 0.024 \\
1017498 & + subject prior (v19) & 9 & 0.176 & 0.169 & 0.164 & 0.006 \\
1017499 & + subject prior (v19) & 9 & 0.177 & 0.159 & 0.151 & 0.008 \\
1017545 & + subject prior (v23) & 8 & 0.214 & 0.195 & 0.183 & 0.011 \\
1017562 & + subject prior (v25) & 8 & 0.195 & 0.179 & 0.169 & 0.010 \\
1017629 & + subject prior (v27) & 9 & 0.203 & 0.184 & 0.167 & 0.017 \\
1017607 & final, with item ratings & 8 & 0.207 & 0.181 & 0.175 & 0.006 \\
1017628 & final, with item ratings & 9 & 0.156 & 0.130 & 0.126 & 0.004 \\
1017682 & final, with item ratings (selected) & 9 & 0.199 & 0.164 & 0.149 & 0.014 \\
\bottomrule
\end{tabular}
\end{table}

\section{The other parts of the predictor}
\label{app:predictor}

\begin{table}[H]
\centering
\caption{The parts of the predictor besides the subject prior. All predictions are clipped to $[0.01, 0.99]$. The code is \texttt{submission/model.py} in the release.}
\label{tab:parts}
\small
\begin{tabular}{p{0.24\linewidth}p{0.68\linewidth}}
\toprule
Part & What it does \\
\midrule
Item ratings & The language model rates each question from 0 to 100 for a typical system. Before the first answer, and without the subject prior, the prediction is $0.5 + 0.25\,(r/100 - 0.5)$ for the rating $r$ clipped to 2--98. Later, a ridge slope on the standardised logit rating shifts the prediction. \\
Other subjects & The rate of the subject is pooled with the rate of the other subjects on the benchmark, with eight pseudo-counts. The subject prior shifts that rate by half of the logit difference that the language model sees between the subjects. Used when other subjects have at least ten answers. \\
Similar questions & Each question adds a shrunk mean of the residuals of its 30 most similar revealed questions, by character 3- to 5-gram cosine. \\
Identical questions & When other subjects answered the same question, one Newton step of a Rasch model shifts its logit. \\
Acquisition & Reveal each streamed question that no other subject has revealed, and reveal all once the stream left is no longer than the budget left. It never uses the outcomes. \\
\bottomrule
\end{tabular}
\end{table}

\end{document}